\documentclass{bmvc2k}
\usepackage{lineno,hyperref}

\usepackage{amssymb}
\usepackage{graphicx}
\usepackage{times}
\usepackage{epsfig}
\usepackage{graphicx}
\usepackage{amsmath}
\usepackage{amssymb}
\usepackage{multirow}
\usepackage[normalem]{ulem}
\usepackage{comment}
\usepackage{xcolor}
\usepackage{setspace}
\usepackage{array}
\usepackage{sidecap}
\usepackage{booktabs}
\usepackage{tabularx} 

\def\@separator{\def\@separator{, }}

\title{Mitigating the Hubness Problem for Zero-Shot Learning of 3D Objects}

\addauthor{Ali Cheraghian$^{1,}$}{ali.cheraghian@anu.edu.au}{2}
\addauthor{Shafin Rahman$^{1,}$}{shafin.rahman@anu.edu.au}{2}
\addauthor{Dylan Campbell}{dylan.campbell@anu.edu.au}{1}
\addauthor{Lars Petersson$^{1,}$}{lars.petersson@data61.csiro.au}{2}

\addinstitution{
 Australian National University,\\ Australia
}
\addinstitution{
Data61 -- CSIRO, Australia
}

\runninghead{A. Cheraghian et. al}{Mitigating the Hubness Problem}

\def\etal{\emph{et al}\bmvaOneDot}

\DeclareMathOperator{\E}{E}
\DeclareMathOperator{\Var}{Var}

\begin{document}

\maketitle

\begin{abstract}

The development of advanced 3D sensors has enabled many objects to be captured in the wild at a large scale, and a 3D object recognition system may therefore encounter many objects for which the system has received no training. Zero-Shot Learning (ZSL) approaches can assist such systems in recognizing previously unseen objects. Applying ZSL to 3D point cloud objects is an emerging topic in the area of 3D vision, however, a significant problem that ZSL often suffers from is the so-called hubness problem, which is when a model is biased to predict only a few particular labels for most of the test instances. We observe that this hubness problem is even more severe for 3D recognition than for 2D recognition. One reason for this is that in 2D one can use pre-trained networks trained on large datasets like ImageNet, which produces high-quality features. However, in the 3D case there are no such large-scale, labelled datasets available for pre-training which means that the extracted 3D features are of poorer quality which, in turn, exacerbates the hubness problem. In this paper, we therefore propose a loss to specifically address the hubness problem. Our proposed method is effective for both Zero-Shot and Generalized Zero-Shot Learning, and we perform extensive evaluations on the challenging datasets ModelNet40, ModelNet10, McGill and SHREC2015. A new state-of-the-art result for both zero-shot tasks in the 3D case is established.

\end{abstract}

\section{Introduction}
\label{sec:intro}

3D point cloud recognition systems have achieved remarkable performance improvements over the past few years. Recent methods employ deep end-to-end learning, sophisticated convolutions that consider the geometric relationship between points, and advanced pooling operators that ensure permutation invariance, resulting in an accuracy of more than 90\% on the popular ModelNet40 dataset \cite{Article2,Article24,Article27,Article28,Article29,Article54}. However, such improvements are limited by the small number of object classes (less than 100). In comparison to 2D recognition, 3D recognition is still in its infancy. Moreover, due to the improvement of new camera sensors, we can easily and cheaply collect more 3D models \cite{rs10020328,Izadi_3D_2011}. In a real-life scenario, we have started encountering many new objects for which a traditional 3D recognition system has not received any training. Hence it is time to investigate Zero-Shot Learning (ZSL) to recognize unseen 3D point cloud objects.

\begin{SCfigure}
\vspace{.1cm}
\hspace{-.4cm}
\includegraphics[width=0.5\linewidth,trim=0cm 0cm 0cm 0cm, clip]{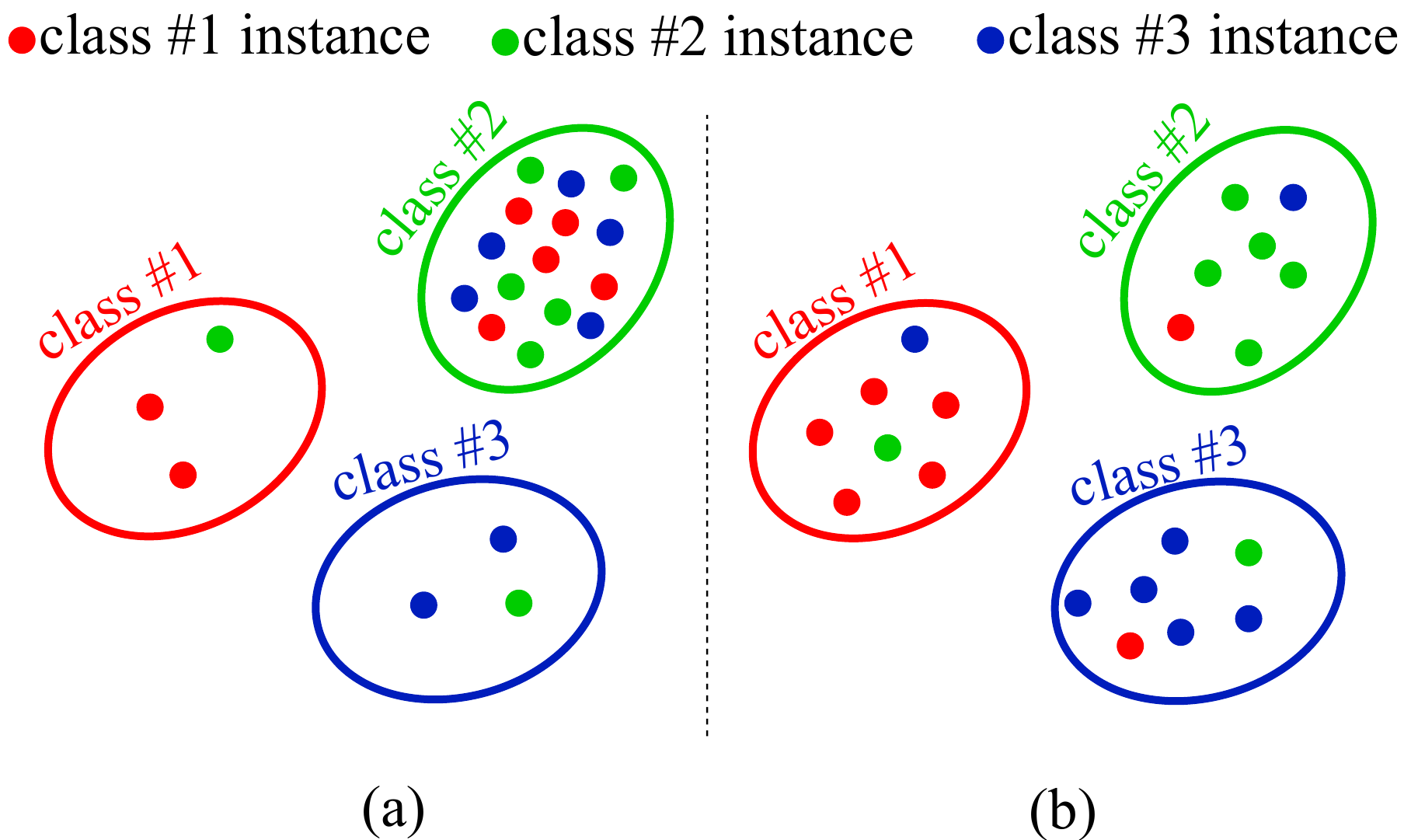}
\caption{\small An illustration of input feature space with (a) severe hubness, and (b) reduced hubness, while performing zero-shot learning of 3D point cloud objects. When the hubness problem exists, most predicted labels come from a small subset of classes (hubs) like {\color{green}class \#2} in (a). By reducing the hubness problem, class prediction is distributed more evenly among classes, as shown in (b). 
The hubness problem becomes more severe for higher dimensions than this toy 2D example.}
\label{fig:skewness}
\end{SCfigure}

ZSL research in computer vision has been largely restricted to 2D images \cite{Hinton_NIPS_2009, Changpinyo_2016_CVPR, Akata_PAMI_2016, Zhang_2017_CVPR, Xian_CVPR_2017}. Moving from 2D images to 3D point clouds for ZSL brings new challenges.
Many deep learning models for 2D images rely on pre-trained deep features that are obtained by considering thousands of classes and millions of images~\cite{Xian_CVPR_2017}. Thus, the extracted 2D features obtained from such pre-trained models tend to be well clustered.
 By contrast, there is no parallel in the 3D point cloud domain; labeled 3D datasets are typically small and have only a few number of classes. For instance, pre-trained models like PointNet \cite{Article1} are trained on ModelNet40~\cite{Article10}, which has only a few thousand instances from 40 classes. This results in poor-quality 3D features with clusters that are not comparable in quality to those obtained from 2D image features.
Therefore, relating those features to their corresponding semantics is more difficult in 3D than 2D, exacerbating the hubness problem~\cite{Zhang_2017_CVPR} of ZSL. The hubness problem occurs when a model gets biased to predict a few particular labels for most of the test instances. In this work, we investigate the hubness problem for ZSL of 3D point cloud objects and propose a new loss to alleviate this problem (see Figure~\ref{fig:skewness}). We calculate this loss by evaluating each training batch in an unsupervised manner, and counting the number of times each seen class gets predicted within a batch. This is used to estimate a measure of hubness: the skewness of the current prediction. We minimize the skewness of each batch to reduce the degree of hubness.

In the 2D ZSL literature, methods are often evaluated on the Generalized ZSL (GZSL) task, which better reflects the real-world problem. To the best of our knowledge, GZSL has not been performed on 3D point cloud objects before. In this paper, we evaluate our proposed loss in both ZSL and GZSL scenarios. Overall, our main contributions of this paper are
\begin{itemize}
   \item a new loss addressing the hubness problem of ZSL is proposed;
   \item an evaluation of the ZSL and GZSL tasks for 3D point cloud classification, for the first time in the literature; and
   \item extensive experiments in ZSL scenarios, establishing state-of-the-art performance on the 3D datasets ModelNet40~\cite{Article10}, ModelNet10~\cite{Article10}, McGill~\cite{Article49}, and SHREC2015~\cite{Article48}.
\end{itemize}



\section{Related Work}

\textbf{Zero-Shot Learning (ZSL) on 2D images:} In the image recognition literature, zero-shot learning has made reasonable progress over the past few years~\cite{rahman2018unified,Zhang_2017_CVPR,Akata_PAMI_2016,Changpinyo_2016_CVPR,Hinton_NIPS_2009,Lampert_PAMI_2014,Xian_CVPR_2017}. The objective of such learning is to be able to recognize objects from unseen classes not used during training. For doing so, semantic information about the class labels in the form of attributes/word vectors is used and methods need to align visual features with respect to semantic vectors. Some methods convert image features to the dimension of semantic vectors \cite{Lampert_PAMI_2014} whereas some other converts semantic features to the visual feature dimension to find a matching score between them \cite{Zhang_2017_CVPR}. Alternatively, some methods attempt to find a latent intermediate space to calculate the matching score \cite{Xian_2016_CVPR}. A few recent works showed that comparing the two domains in the visual space is more effective at addressing the hubness problem of zero-shot learning \cite{Zhang_2017_CVPR}. Some other notable research directions in this line of investigation include exploring class attribute association~\cite{Demirel_2017_ICCV}, domain adaptation~\cite{Deutsch_2017_CVPR}, inductive vs. transductive ZSL ~\cite{Kodirov_2017_CVPR, Li_2017_CVPR}, multi-label ZSL~\cite{Lee_2018_CVPR,rahman2018deep}, zero-shot detection~\cite{rahman2018ZSD} etc. In this paper, we apply zero-shot learning on 3D point cloud objects instead of the traditional 2D image.

\noindent
\textbf{Generalized Zero-Shot Learning (GZSL) on 2D images:} While ZSL only focus on recognizing unseen classes during inference time, GZSL attempts to classify both seen and unseen classes together \cite{rahman2018unified,Xian_CVPR_2017,Chao_ECCV_2016}. It means, given an image, GZSL assigns a winning label based on the maximum probability score from the combined seen and unseen probabilities. As the training only observes seen data, the prediction probability of seen classes is usually higher than unseen ones which make seen and unseen scores incompatible. To solve this problem, some methods increase the unseen scores by a constant amount \cite{Chao_ECCV_2016}, some other incorporate a separate training based on the ratio of seen/unseen to balance the probabilities~\cite{rahman2018unified}. Another way of addressing the same problem is to explore a transductive setting where models can take advantage of unlabeled unseen data~\cite{Zhao_NIPS_2018}. In this paper, we also investigate and report the GZSL performance for 3D point cloud objects.


\noindent
\textbf{Zero-shot learning on 3D point clouds:} Recently, there has been significant progress on 3D point cloud classification using deep learning~\cite{Article1,Article2,Article24,Article27,Article28,Article29}. However, to the best of our knowledge, there is only one work~\cite{cheraghian2019zeroshot} that has addressed the ZSL problem for 3D point clouds. In this work, the PointNet~\cite{Article1} architecture is utilized to extract a feature space, and a bilinear compatibility function is applied to associate the point cloud feature vector with the corresponding semantic feature vector.
While it is a pioneering work~\cite{cheraghian2019zeroshot}, they employ the feature space as the embedding space which causes more hubness. They also do not report any results on the GZSL task.
In this paper, we specifically set out to investigate the tasks of ZSL and GZSL for 3D point cloud classification.

\noindent
\textbf{The hubness problem:} The hubness problem in high dimensional nearest neighbor search spaces was first investigated in \cite{Article57} where they illustrate that the hubness problem is related to the data distribution in the high dimensional space. In recent studies \cite{article56,Shigeto_Hubness_2015,Zhang_2017_CVPR}, the hubness problem in ZSL is investigated. Dinu \etal~\cite{article56} proposed an algorithm that corrects the hubness problem by using more unlabeled seen data in addition to test instances.
Shigeto \etal~\cite{Shigeto_Hubness_2015} mentioned that the projection function used for least squares regularization effect the hubness problem negatively and instead introduces a reverse regularized function in order to weaken the hubness problem. In contrast to the mentioned works, Zhang \etal~\cite{Zhang_2017_CVPR} proposed to deal with the hubness problem by instead considering the feature space as the embedding space. To the best of our knowledge, there is no previous work that addresses the hubness problem of ZSL on 3D point cloud classification.







\section{The Hubness Problem}

The hubness problem is related to the curse of dimensionality associated with nearest neighbor (NN) search~\cite{Article57}. That is, in high-dimensional data some points, called hubs, frequently occur in the $k$-nearest neighbor set of other points. In ZSL, the hubness problems occurs for two reasons \cite{Shigeto_Hubness_2015}. Firstly, both input and semantic features reside in a high-dimensional space. Secondly, ridge regression, which is widely used in ZSL, is known to induce hubness.
As a result, it causes a bias in the predictions, with only a few classes predicted most of the time regardless of the query. To calculate the degree of hubness in a nearest neighbor search problem, the skewness of the empirical distribution $\rho _{j}$ can be used \cite{Shigeto_Hubness_2015,Article57}. The distribution $\rho_{j}$ counts the number of times ($\rho_{j}(i)$) the $i$\textsuperscript{th} point (known as the prototype) is in the top $j$ nearest neighbors of the test samples. The skewness of this distribution is defined as
\begin{align}
\rho_{j}\textrm{-skewness}=\frac{\sum_{i=1}^{n}(\rho_{j}(i)-\E\left [ \rho_{j} \right ])^3}{n \left( \Var\left [ \rho_{j} \right ] \right)^{\!\frac{3}{2}}}
\end{align}
where $n$ is the number of test prototypes. Large values of skewness indicate that the feature space is severely affected by the hubness problem.

\begin{SCfigure} 
\includegraphics[width=.48\linewidth,trim=.05cm -1.2cm .05cm .0cm, clip]{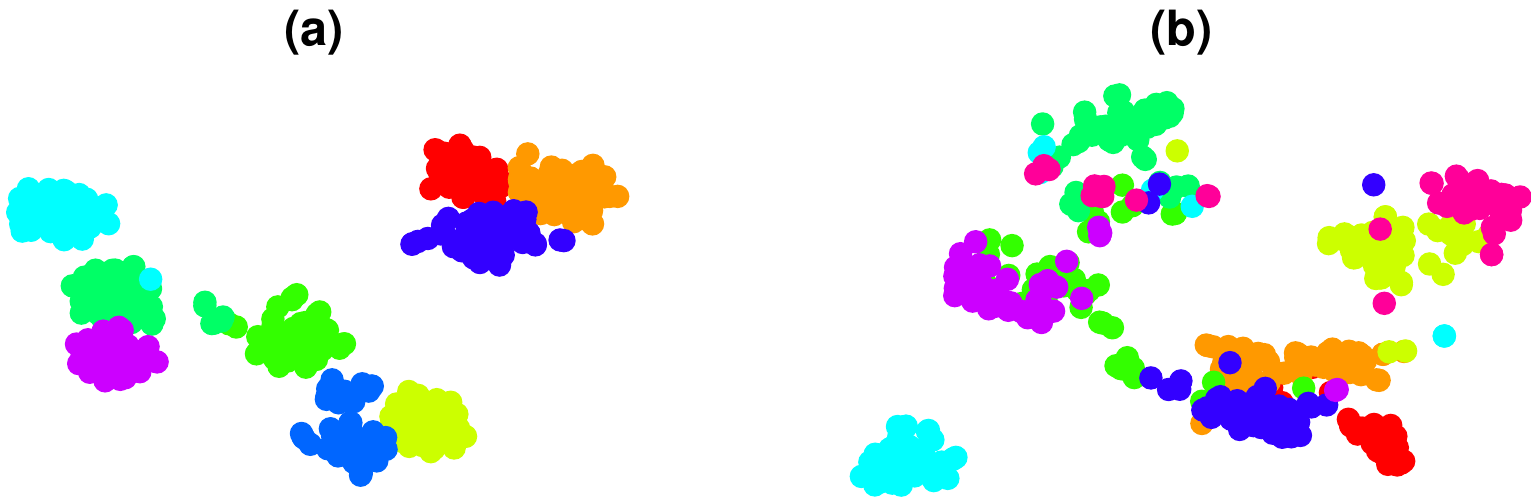}
\caption{\small tSNE~\cite{tSNE_van2014} visualization of (a) 2D image features of 9 randomly-selected classes of the CUB~\cite{CUB_2011} dataset, and (b) 3D point cloud features of the ModelNet10~\cite{Article10} dataset. The cluster structure in the 2D feature space is much better defined, with tighter and more separated clusters than those in the 3D point cloud feature space.}
\label{fig:visual_feature}
\end{SCfigure}

We observe empirically that the hubness problem is more acute in the feature space of 3D point clouds than in the feature space of 2D images. Some intuition for this can be attained by visualizing the respective pre-trained feature spaces, as shown in Figure~\ref{fig:visual_feature}, for 500 instances of \textasciitilde9 classes from the 2D CUB~\cite{CUB_2011} dataset~(a) and the 3D ModelNet10~\cite{Article10} dataset~(b). The quality of the image features is much higher than the point cloud features, with a much more separable cluster structure. When the clusters are not well-separated, the hubness problem is worsened. To address the hubness problem in 3D point cloud ZSL, we introduce a novel loss which reduces hubness in an unsupervised manner using only seen instances during training.

\section{Method}
\subsection{Problem Formulation}

Let a 3-dimensional point cloud with $n$ points $\mathbf{x}$ be defined as $\mathcal{X}=\left \{ \mathbf{x}_{1},...,\mathbf{x}_{n} \right \}
$. We are given seen $\mathcal{T}^{s}$ and unseen $\mathcal{T}^{u}$ class label sets, consisting of $p$ and $q$ class labels respectively, with $\mathcal{T}^{s} \cap \mathcal{T}^{u}=\varnothing$. We also have a set of $d$-dimensional semantic representations for each class label in both the seen and unseen sets, denoted as $\mathcal{E}^{s} $ and $\mathcal{E}^{u} $ respectively. Then, we define a seen set as $\mathcal{D}^{s}=\left \{ (\mathcal{X}^{s}_{i},t^{s}_{i},\mathbf{e}^{s}_{i}): i = 1,...,n_{s} \right \}$, where $\mathcal{X}^{s}_{i}$ is the $i$\textsuperscript{th} point cloud of the seen set with the label $t_{i}^{s} \in \mathcal{T}^{s}$ and its associated semantic representation $\mathbf{e}^{s}_{i} \in \mathcal{E}^{s}$, and $n_{s}$ is the number of seen instances. Similarly, the set of unseen instances is defined as $\mathcal{D}^{u}=\left \{ (\mathcal{X}^{u}_{i},t^{u}_{i},\mathbf{e}^{u}_{i}): i = 1,...,n_{u} \right \}$, where $\mathcal{X}^{u}_{i}$ is the $i$\textsuperscript{th} point cloud of the unseen set with the label $t_{i}^{u} \in \mathcal{T}^{u}$ and its associated semantic representation $\mathbf{e}^{u}_{i}\in \mathcal{E}^{u}$, and $n_{u}$ is the number of unseen instances. To place the problem in a zero-shot setting, it is crucial to state that $\mathcal{D}^{u}$, $\mathcal{T}^{u}$ and $\mathcal{E}^{u}$ are not observed during the training stage. Here, we define the ZSL and GZSL tasks addressed in this paper:
\vspace{-2.5mm}
\begin{itemize}
   \item \textbf{Zero Shot Learning (ZSL):} The task of assigning an unseen class label $\hat{t}^{u} \in \mathcal{T}^{u}$ to a given unseen 3D point cloud $\mathcal{X}^{u}$. 
   \vspace{-2mm}
   \item\textbf{Generalized Zero Shot Learning (GZSL):} The task of assigning a class label $\hat{t} \in \mathcal{T}^{s}\cup \mathcal{T}^{u}$, which can belong to either the seen and unseen classes, to a given 3D point cloud $\mathcal{X} \in \mathcal{D}^{s} \cup \mathcal{D}^{u} $. 
\end{itemize}

\subsection{Training}

\begin{SCfigure} 
\includegraphics[width=.45\linewidth,trim=0cm 0cm 0cm .0cm, clip]{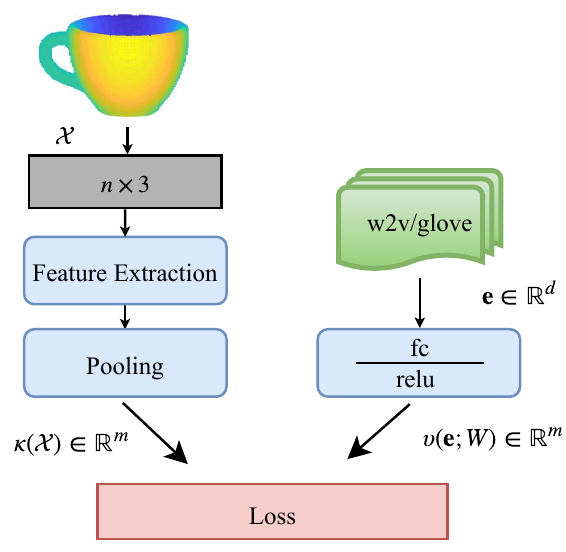}
\caption{\small Illustration of the model architecture of the proposed approach. The point cloud encoding network is shown, consisting of a feature extraction module to extract a rich representation of the point cloud, and a pooling module to ensure the permutation invariance of the point cloud features. 
It takes a point cloud $\mathcal{X}$ as input, and outputs an $m$-dimensional feature vector $\mathbf{\kappa (\mathcal{X})} \in \mathbb{R}^{m}$, which is the point cloud feature. In the other branch, the semantic feature vector $\mathbf{e}$ is projected to the point cloud feature space \cite{Zhang_2017_CVPR} using three fully-connected layers, and outputs an ${m}$-dimensional semantic embedding vector $\upsilon (\mathbf{e};W) \in \mathbf{R}^{{m}}$.}
\label{fig:artitecture}
\end{SCfigure}

The architecture of the proposed model is shown in Figure \ref{fig:artitecture}. The architecture consists of two branches. In the left branch, a feature vector 
$\kappa(\mathcal{X})\in \mathbb{R}^{m}$ 
from a point cloud $\mathcal{X}$ is extracted using a point cloud network, which can be any network that learns a feature space from 3D point cloud data with the capability of being invariant to permutations in the point cloud \cite{Article1,Article2,Article24,Article27,Article28,Article29}. In the right branch, a semantic feature vector $\mathbf{e}\in \mathbb{R}^{d}$ is mapped into point cloud feature space using a semantic projection network $\upsilon(\cdot)$, which consists of three fully-connected layers with $512$, $768$, and $1024$ dimensions respectively and trainable weights $W$, each followed by a ReLU nonlinearity.


In order to train the proposed architecture, a loss function   
\begin{align}
L_{T} = L_{S} + \alpha L_{U} 
\label{Equation}
\end{align}
is minimized, where $L_{S}$ is a supervised distance loss that minimizes dissimilarity between the projected semantic representation and point cloud features using the ground truth seen, labeled data. This term aligns the semantic vectors and point cloud feature vectors to each other. The term  $L_{U}$ is an unsupervised skewness loss as it does not use ground truth and quantifies the severity of hubness in the model. In general, minimizing the $L_{T}$ loss learns a latent space where dissimilarity between visual features and semantic features is minimized, and the hubness problem is reduced. The combination weight $\alpha$ is used to balance the $L_{S}$ and $L_{U}$ losses. Both loss terms will be discussed in detail below in two separate sections, "supervised distance loss" and "unsupervised skewness loss".


\noindent
\textbf{Supervised distance loss:} The supervised distance loss $L_{S}$ considers labels of the seen instances and encourages semantic vectors to align with the point cloud feature vectors, such that each semantic representation related to a seen class is mapped to the cluster of feature vectors associated with that seen class. $L_{S}$ is defined as 
\begin{align}
L_{S} = \frac{1}{N}\sum_{i=1}^{N}\left \| \kappa(\mathcal{X}^{s}_{i}) - \upsilon(\mathbf{e}^{s}_{i};W) \right \|^2_{2}  + \lambda \left \| W \right \|_{2}^{2}
\end{align}
where $N$ is the number of samples in the batch, $\kappa(\mathcal{X}_{i}^{s}) \in \mathbb{R}^{m}$ is the point cloud feature vector associated with point cloud $\mathcal{X}_{i}^{s}$, $W$ are the weights of the nonlinear projection function $\upsilon(\cdot)$ that maps from the semantic embedding space $\mathcal{E}$ to the point cloud feature space, and $\lambda$ is used to control the effect of regularization loss to the total loss.

\noindent
\textbf{Unsupervised skewness loss:} The label predicted for the $i$\textsuperscript{th} seen instance in a batch with size $N$ is defined as 
\begin{align}
\hat{t}^{s}_{i}= \underset{t\in \mathcal{T}^{s}}{\arg \max} \,\cos(\kappa(\mathcal{X}^{s}),\upsilon({\mathbf{e}(t)};W))
\end{align}
where $\mathbf{e}(t)$ is the semantic vector associated with label $t \in \mathcal{T}^{s}$. Then for all instances in the batch, we predict their labels, and define a set $\hat{\mathcal{T}}^{s}= {\left \{ \hat t^{s}_{1},..., \hat t^{s}_{N} \right \}}$. We then calculate the frequency of each class from $\hat{\mathcal{T}}^{s}$ by using the histogram function $\mathcal{H}(\hat{t}^{s}_{i})$, which counts the number of times that a specific seen class is predicted. This function has the property that $\sum_{i=1}^{p}\mathcal{H}(\hat{t}^{s}_{i})=N$. Finally, we define the skewness loss as
\begin{align}
L_{U} = \frac{1}{N (\Var[\mathcal{H}(\hat{\mathcal{T}}^{s})])^{\frac{3}{2}}}\sum_{i=1}^{N}(\mathcal{H}(\hat{t}^{s}_{i})-\E[\mathcal{H}(\hat{\mathcal{T}}^{s})])^{3}
\label{eq:skewloss}
\end{align}
where $\mathcal{H}(\hat{t}^{s})$ represents the statistics of prediction for all instances, that is, how many times each output is predicted regardless of being true or false. 

\subsection{Inference}

During training, which uses the seen instances $\mathcal {X}^{s}$, the distance between point cloud feature vectors and the projected semantic representation is minimized such that it can classify a seen 3D point cloud $\mathcal{X}^s$ as belonging to a certain seen class label $t \in \mathcal{T}^{s}$ by finding the nearest neighbor. A similar method can be used to classify an unseen 3D point cloud $\mathcal{X}^u$ belonging to an unseen class, by predicting
\begin{align}
\hat{t}^{u}= \underset{t\in \mathcal{T}^{u}}{\arg \max} \,\cos(\kappa(\mathcal{X}^{u}),\upsilon(\mathbf{e}(t);W))
\end{align}
where $\mathbf{e}(t)$ is the semantic vector associated with label $t \in \mathcal{T}^{u}$.

For the GZSL task, we follow the method proposed by Chao \etal~\cite{Chao_ECCV_2016}.
They observe that the performance of unseen classes drops significantly in GZSL, where the label space is jointly seen and unseen classes, when compared to conventional ZSL, while the performance of the seen classes is almost the same as for the multi-class task. Since the network is trained using only seen classes during training, the scores of seen classes are greater than those of unseen classes. As a result, the network tends to predict seen classes, even when the test sample comes from an unseen class. This leads to a signficant performance drop in GZSL. To alleviate this problem, a weight factor $\beta$ is subtracted from the seen prediction output to reduce their effect on the overall prediction, where $\beta$ is calculated by Monte Carlo cross-validation~\cite{XU20011}. Hence for GZSL we predict the label
\begin{align}
\hat{t}= \underset{t\in \mathcal{T}^{s} \cup \mathcal{T}^{u} }{\arg \max} \,\cos\big(\kappa(\mathcal{X}),\upsilon(\mathbf{e}(t);W))- \beta \mathbb{I}\left [ t\in\mathcal{T}^{s} \right ]\big)
\end{align}

\noindent
where $\mathbb{I}\left [ \cdot  \right ]\in\left \{ 0,1 \right \}$ indicates if $t$ is a seen class or not, and $\mathbf{e}(t)$ is the semantic vector associated with label $t \in \mathcal{T}^{u} \cup  \mathcal{T}^{u}$.

\section{Experiments}

\subsection{Setup}

\noindent
\textbf{Dataset:} In this paper, four 3D datasets, ModelNet40~\cite{Article10}, ModelNet10~\cite{Article10}, McGill~\cite{Article49}, and SHREC2015~\cite{Article48}, are used to evaluate our proposed method. Also one 2D dataset, CUB~\cite{CUB_2011}, is used to evaluate our proposed method. The dataset statistics are shown in Table~\ref{Table:splitting}. We use the split protocol proposed by Cheraghian \etal~\cite{cheraghian2019zeroshot}, where the selected set of seen classes from ModelNet40~\cite{Article10} are determined based on those which do not belong to the related ModelNet10 \cite{Article10} dataset. As a result, 30 classes from ModelNet40 \cite{Article10} are selected as seen classes. The unseen set is sourced from three datasets with different classes, ModelNet10~\cite{Article10}, McGill~\cite{Article49} and SHREC2015~\cite{Article48}. For the 2D datasets, we follow the Standard Splits (SS) of Xian \etal~\cite{Xian_CVPR_2017}.

\begin{SCtable}
  \centering \setlength\tabcolsep{2.5pt}

\scalebox{.85}{

\begin{tabular}{llcccc}
\hline
&\multirow{2}{*}{Dataset}    & total&seen/ & total & training/ \\
&    & class&unseen & models & valid/testing \\ 
    \hline
\multirow{4}{*}{3D}&ModelNet40 \cite{Article10} & 40&30/-        & 12,311        & 5852/1560/--      \\ 
&ModelNet10 \cite{Article10} & 10&-/10        & 4,899        & --/--/908     \\ 
&McGill \cite{Article49}    & 19&-/14        & 456        & --/--/115     \\ 
&SHREC2015 \cite{Article48}  & 50&-/30        & 1,200        & --/--/192     \\ \hline

2D&CUB \cite{CUB_2011}& 200&150/50        & 11788        & 8855/--/2933     \\ \hline
\end{tabular}}
\caption{\small Dataset statistics. For 3D experiments, seen classes from ModelNet40 are used during training. Unseen classes taken from ModelNet10, McGill and SHREC2015 are used during evaluation. For 2D image related experiments, we use standard settings of CUB dataset.}
\label{Table:splitting}
\end{SCtable}

\noindent
 \noindent
\textbf{Semantic features:} ZSL methods on 2D image data are often evaluated using both supervised (attributes) and unsupervised (word vector) embeddings. Many image datasets, e.g., Animals with Attributes (AwA)~\cite{AwA_2009}, and Caltech-UCSD Birds (CUB)~\cite{CUB_2011}, come with attribute annotations. However, 3D point cloud datasets, to date, do not contain such attributes. Therefore, here, we work with unsupervised word vectors obtained from an unannotated text corpus. We have used  $\ell_2$ normalized, 300 dimensional word2vec~\cite{Mikolov_NIPS_2013} and GloVe~\cite{Jeffrey_GloVe_2014} word vectors. Also, the 312-dimensional attribute vectors from Wah \etal~\cite{CUB_2011} are used for the CUB experiments.



\noindent
\textbf{Evaluation:} In this work, the recognition performance is measured by top-$1$ accuracy, which means the class with the highest predicted probability should match the true class to be considered ``correct". Also, the performance of GZSL is calculated by Harmonic Mean (HM)~\cite{Xian_CVPR_2017}.




\noindent
\textbf{Cross Validation:} In this paper, Monte Carlo cross-validation \cite{XU20011} was used to find the best hyper-parameters. This process was repeated 10 times, and parameters were calculated by taking the average. The hyper-parameters batch size, $\alpha$, $\lambda$, and $\beta$ were 64, 0.7, 0.0001, 0.6 respectively.


\noindent
\textbf{Implementation details\footnote{We will release the code and data when published.}:} We used the following set-up during training in all of the experiments. We used the Adam optimizer~\cite{Article40} with an initial learning rate of 0.001 and a batch size of 64. For the point cloud network, we used PointNet~\cite{Article1} with five shared mlp layers (64,64,64,128,1024) followed by a max pooling layer and two fully connected layers (512,1024), which resulted in a 1024-dimensional feature vector as the input embedding feature. Batch  normalization  (BN)  ~\cite{Article45} and  ReLU  functions  were  used  for each layer. PointNet was pre-trained on the 30 seen classes of ModelNet40. For the semantic projection layers, we used three fully-learnable fully-connected layers (512,768,1024) with ReLu non-linearities. We implemented the architecture using TensorFlow~\cite{Article46}. 

\begin{table}[!t]
    
    \begin{subtable}
      \centering
      \scalebox{.73}{
\begin{tabular}{lcc|cc|cc}
\hline
\multirow{2}{*}{Method} & \multicolumn{2}{c|}{ModeNet10} & \multicolumn{2}{c|}{McGill} & \multicolumn{2}{c}{SHREC2015} \\
 & w2v & GloVe & w2v & GloVe & w2v & GloVe \\ \hline
 \cite{cheraghian2019zeroshot} & \multicolumn{1}{c|}{27.0} & 14.8 & \multicolumn{1}{c|}{9.8} & 7.2 & \multicolumn{1}{c|}{4.1} & 3.6 \\
baseline & \multicolumn{1}{c|}{30.6} & 25.1 & \multicolumn{1}{c|}{10.7} & 9.8 & \multicolumn{1}{c|}{4.1} & 4.1 \\ \hline
Ours & \multicolumn{1}{c|}{\textbf{33.9}} & \textbf{28.7} & \multicolumn{1}{c|}{\textbf{12.5}} & \textbf{11.1} & \multicolumn{1}{c|}{\textbf{6.2}} & \textbf{4.2} \\ \hline
\end{tabular}}
    \end{subtable}%
    \begin{subtable}
      \centering
        \scalebox{.73}{
\begin{tabular}{lccc|ccc}
\hline
\multirow{2}{*}{Method} & \multicolumn{3}{c|}{w2v} & \multicolumn{3}{c}{GloVe} \\
 & seen & unseen & HM & seen & unseen & HM \\ \hline
\cite{cheraghian2019zeroshot} & 40.1 & 22.5 & 28.8 & 49.2 & 18.2 & 26.6 \\

Baseline & 43.6 & 26.2 & 32.7 & 52.4 & 21.0 & 30.0 \\ \hline
Ours & \textbf{53.8} & \textbf{26.2} & \textbf{35.2} & \textbf{53.8} & \textbf{25.7} & \textbf{34.8} \\ \hline
\end{tabular}}
    \end{subtable} 
    \vspace{-8pt}
    \caption{\small (Left) Overall performance of ZSL using the ModelNet10, McGill and SHREC2015 dataset. (Right) GZSL performance on ModelNet10.}
    
\label{fig:ZSL_GZSL}

\end{table}

\begin{figure} 
\centering
\includegraphics[width=1\linewidth,trim=0cm 0cm 0cm .0cm, clip]{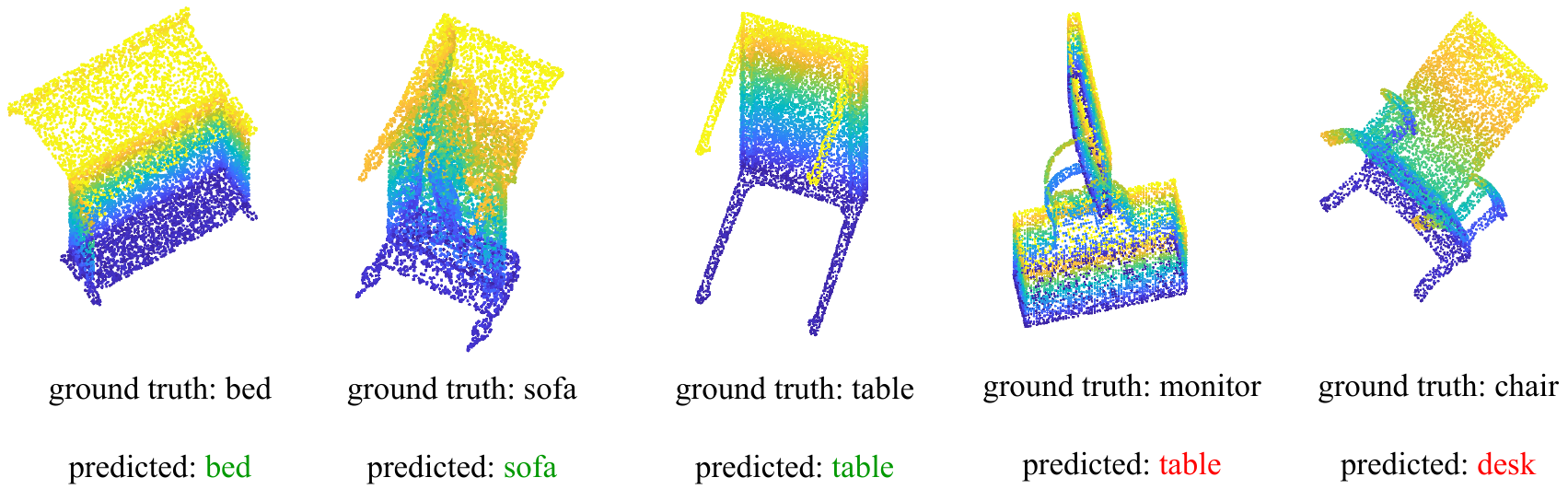}
\caption{\small Qualitative results of successful (green) and failed (red) cases of our proposed ZSL method.}

\label{fig:visualization}
\end{figure}

\subsection{Overall Results}

For both ZSL and GZSL, we compare our proposed method with the work of Cheraghian \etal~\cite{cheraghian2019zeroshot} as well as a baseline. Cheraghian \etal has only reported on the task of ZSL, but for the sake of comparison we implement their method on the GZSL setting as well. In the baseline method, we use only the supervised distance loss, $L_{S}$. 

\noindent
\textbf{ZSL results:} In this subsection, the top-1 accuracy performance of three different structures, Cheraghian \etal \cite{cheraghian2019zeroshot}, the baseline and ours, based on the two semantic representations, {\it w2v} and {\it GloVe}, are shown. For  the {\it w2v} semantic representation, as shown in Table~\ref{fig:ZSL_GZSL}, the winning architecture for point cloud zero-shot learning is our proposed method which achieves an accuracy of 33.9\%, 12.5\%, and 6.2\% on the ModelNet10, McGill, and SHREC2015 datasets respectively. Also, for the {\it w2v} semantic representation, the winning architecture for point cloud zero-shot learning is our proposed method which achieves an accuracy of 28.7\%, 11.1\%, and 4.2\% for the ModelNet10, McGill, and SHREC2015 datasets respectively. We make the following observations: 1) {\it w2v}, in comparison to GloVe, performs better on the ZSL task which may be a result of point cloud data being more alignable to {\it w2v} semantics than {\it GloVe}. As shown in this table, projecting the semantic space to the feature space, which is done in the baseline and our proposed methods, achieves a superior performance over that of projecting the feature space to the semantic space as in Cheraghian \etal \cite{cheraghian2019zeroshot}. This is due to the observation that the hubness problem is more likely to arise in the semantic space than the feature space \cite{Zhang_2017_CVPR}. We have presented some qualitative visualization of successful and failed cases in Figure \ref{fig:visualization}.

\noindent
\textbf{GZSL results:} GZSL is a significantly more difficult task than ZSL. It is also closer to a realistic scenario, as in the ZSL scenario only unseen classes are considered during testing, whereas both seen and unseen instances are considered in the GZSL setting. 
Usually, methods proposed for the ZSL task do not report on GZSL. However, in this paper we report GZSL, to the best of our knowledge, for the first time on 3D object recognition. The obtained result is shown in Table~\ref{fig:ZSL_GZSL}. As shown, seen and unseen, and HM accuracy are improved using our proposed method for both {\it w2v} semantics and {\it GloVe}.


\begin{SCfigure} 
\includegraphics[width=.48\linewidth,trim=0cm 0cm 0cm 0.cm, clip]{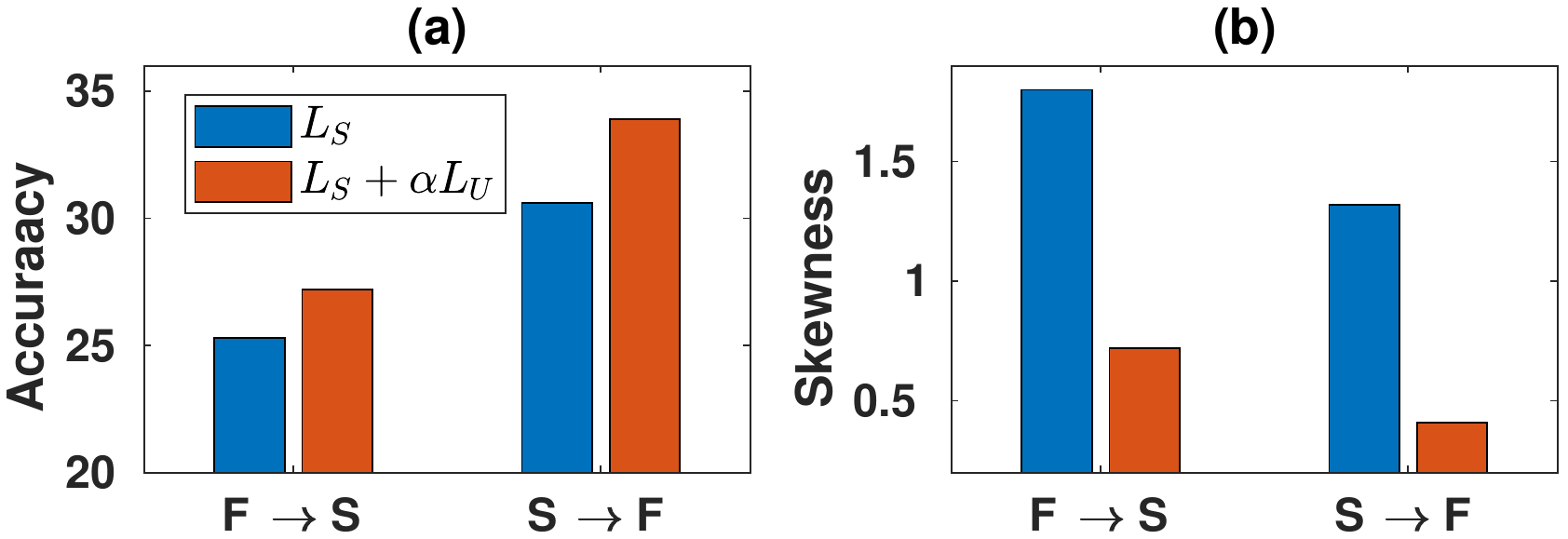}
\caption{\small Effects of selecting different embeddings on: (a) accuracy (b) skewness value as hubness measure while appying ZSL on 3D point cloud objects. $F$ and $S$ represent the point cloud feature and the semantic embedding respectively.}

\label{fig:ablation_study}
\end{SCfigure}
\subsection{Ablation study}

In this section, we evaluate the importance of selecting the appropriate embedding space for the ZSL task. As shown in Figure~\ref{fig:ablation_study}~(a), using the skewness loss, $L_{U}$, is effective in both "feature to semantic" and "semantic to feature" projections. As discussed in \cite{Zhang_2017_CVPR}, in order to reduce the hubness problem, the skewness needs to be made as small as possible. Therefore, we proposed the skewness loss, and its effect on skewness of the seen classes is shown in Figure~\ref{fig:ablation_study}~(b). The skewness value drops after applying our skewness loss in both the "feature to semantic" and "semantic to feature" projections.

\subsection{Experiments Beyond 3D}
In Table \ref{table:ZSL_2D}, we report the performance when applying our proposed skewness loss to the more common problem of 2D image data of the CUB dataset, using attributes as the semantic information. Our result outperforms the current state-of-the-art methods of ZSL. Our closest competitor, Zhang et al. \cite{Zhang_2017_CVPR}, also addressed the hubness problem, however, our method explicitly minimizes the hubness measure which turns out to be more effective.


\begin{table}[]
\centering
\begin{tabular}{lcccccc}
\hline
Method & SJE~\cite{Akata_CVPR_2015} & ESZSL~\cite{romera_ICML_2015} & SynC~\cite{Changpinyo_2016_CVPR} & SEC~\cite{bucher_ECCV_2016} & DEM~ \cite{Zhang_2017_CVPR} & Ours \\ \hline
Accuracy & 50.1 & 47.2 & 54.5 & 43.3 & 58.3 & \textbf{59.4} \\ \hline
\end{tabular}
\caption{\small Experiments on the CUB image dataset}
\label{table:ZSL_2D} 
\end{table}


\section{Conclusion}
With the aid of better 3D capture systems, obtaining 3D point cloud data of objects at a very large scale has become more feasible than before. However, 3D point cloud recognition systems have not scaled up to handle this large scale scenario. To readjust such a system with newly available data that have not observed during training, we apply a zero-shot learning approach to facilitate classification of previously unseen input. Similar to ZSL on 2D images, we notice that such classification of 3D point clouds suffers from the hubness problem. Moreover, the hubness problem in 3D is more severe than that observed in the 2D case. One possible reason could be that the 3D features are not trained on millions of 3D instances in the same way that 2D convolutional networks can be. In this paper, we attempt to reduce the effect of the hubness problem while performing ZSL on 3D point cloud objects by proposing a novel loss. In addition, we report results on Generalized ZSL in conjunction with ZSL. Rigorous experiments on both 3D point clouds and 2D image datasets show significant improvement in performance over the current state-of-the-art methods.


{\small
\bibliography{egbib}
}
\end{document}